\documentclass[10pt,twocolumn,letterpaper]{article}

\usepackage{iccv}
\usepackage{times}
\usepackage{epsfig}
\usepackage{graphicx}
\usepackage{amsmath}
\usepackage{amssymb}

\usepackage{subcaption}

\usepackage{lipsum}

\def\ov2slam{OV$^{2}$SLAM}

\DeclareMathOperator*{\argmin}{arg\,min}

\DeclareMathOperator*{\SE3}{\mathbb{SE}(3)}

\usepackage[pagebackref=true,breaklinks=true,letterpaper=true,colorlinks,bookmarks=false]{hyperref}

\iccvfinalcopy 

\def\iccvPaperID{18} 

\ificcvfinal\fi

\begin{document}

\title{Hyperspectral 3D Mapping of Underwater Environments}

\author{Maxime Ferrera $^{1} \;\;$
Aurélien Arnaubec $^{1} \;\;$
Klemen Isteni\v{c} $^{2} \;\;$
Nuno Gracias $^{2} \;\;$
Touria Bajjouk $^{1}$ \\
{\normalsize $^{1}$ French Research Institute for Exploitation of the Sea (IFREMER), France} \\
{\normalsize $^{2}$ Research Centre in Underwater Robotics (CIRS), 
University of Girona, Spain} \\
{\tt\small Email: maxime.ferrera@ifremer.fr}
}


\ificcvfinal\thispagestyle{empty}\fi

\twocolumn[{%
\renewcommand\twocolumn[1][]{#1}%

\maketitle

\ificcvfinal\thispagestyle{empty}\fi

\begin{center}
    \centering
    \captionsetup{type=figure}
    \includegraphics[width=0.85\textwidth]{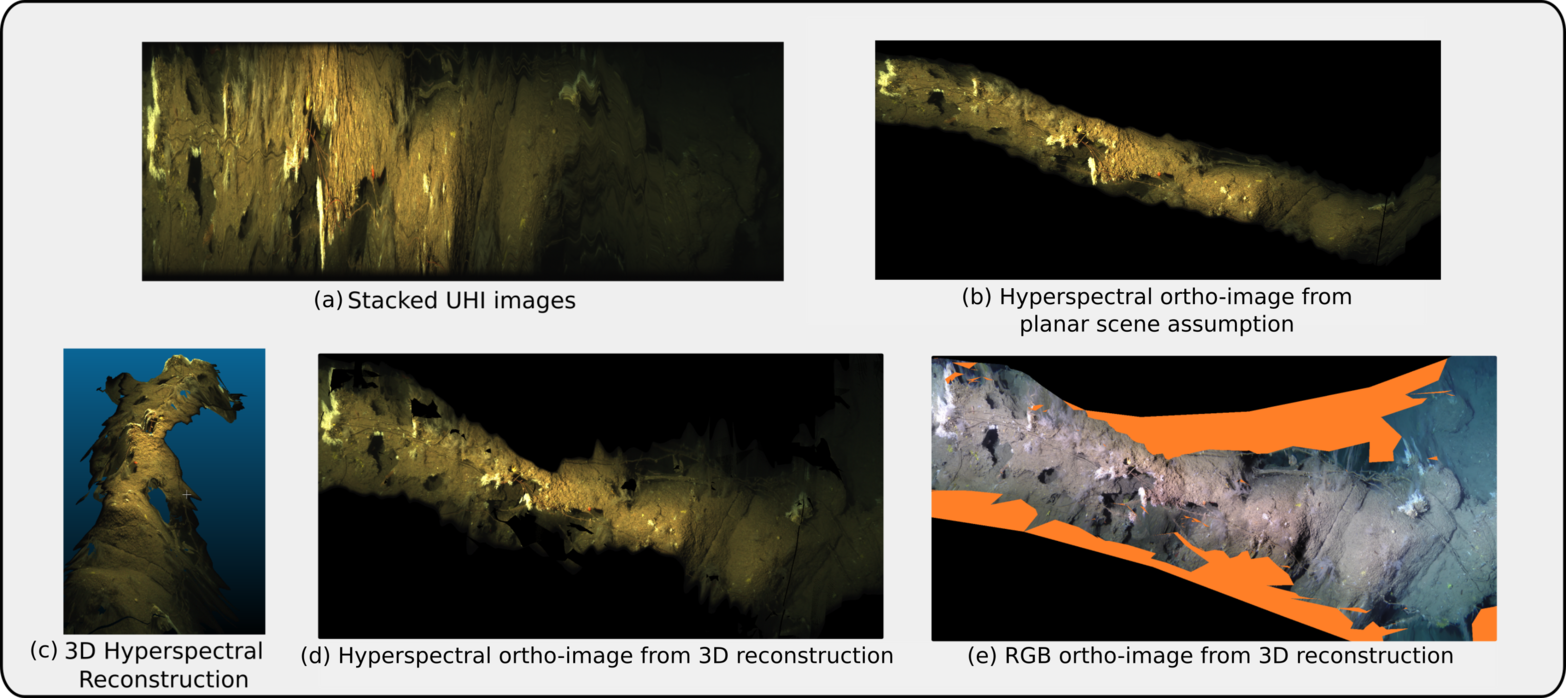}
	\caption{From a collection of single line hyperspectral images captured with an Underwater Hyperspectral Imaging (UHI) push-broom camera (a), the proposed method allows to create accurate hyperspectral 3D models of underwater environments (c) from which geometrically correct ortho-images can be extracted (d).  In opposition to classical hyperspectral mosaicking methods (b), ours is not based on a planar scene assumption and leverage on multi-view 3D reconstruction techniques to produce reliable material for scientific analysis.  Comparing the RGB ortho-image (e) to the hyperspectral one (d) highlights the enhancement of the proposed method over basic mosaicking (b).  Note that for hyperspectral images, RGB-like colors have been simulated here using three spectral bands but hundreds are available, allowing much finer spectral analyses of the data compared to classical RGB cameras.}
	\label{fig:intro}
\end{center}%
}]


\begin{abstract}
    Hyperspectral imaging has been increasingly used for underwater survey applications over the past years. As many hyperspectral cameras work as push-broom scanners, their use is usually limited to the creation of photo-mosaics based on a flat surface approximation and by interpolating the camera pose from dead-reckoning navigation.  Yet, because of drift in the navigation and the mostly wrong flat surface assumption, the quality of the obtained photo-mosaics is often too low to support adequate analysis.
    In this paper we present an initial method for creating hyperspectral 3D reconstructions of underwater environments.  By fusing the data gathered by a classical RGB camera, an inertial navigation system and a hyperspectral push-broom camera, we show that the proposed method creates highly accurate 3D reconstructions with hyperspectral textures.  We propose to combine techniques from simultaneous localization and mapping, structure-from-motion and 3D reconstruction and advantageously use them to create 3D models with hyperspectral texture, allowing us to overcome the flat surface assumption and the classical limitation of dead-reckoning navigation.
\end{abstract}

\section{Introduction}

In coastal areas, the methods of acquiring and analyzing hyperspectral data of shallow waters have been successfully developed in recent years \cite{kutser2020remote}. Thanks to the spectral richness of these sensors, which can capture hundreds of contiguous spectral bands across a wide light spectrum \cite{chennu2017diver}, it is possible to estimate the relative contribution of several properties characterizing the complexity of marine ecosystems.  Hyperspectral cameras deployed both in situ and on airborne platforms, have thus demonstrated their ability to (i) estimate the optical properties of the water column by inversion methods \cite{lee1998hyperspectral,petit2017hyperspectral,jay2017hyperspectral} and (ii) to characterize benthic habitats \cite{chennu2013hyperspectral,vahtmae2020much}. Spatial approaches have also been developed to map and quantify changes in the health status of coral reefs \cite{bajjouk2019detection}. 

Until recently, hyperspectral imaging data has been acquired mostly using passive sensors installed either on aerial platforms or satellites. These sensors use the sun as the light source and record reflected solar radiation. This approach becomes limiting where sunlight is at best partially filtered, if not totally absent, and is thus not applicable for deep underwater environments.  Hence, hyperspectral sensors have been deployed on underwater ROVs platforms to map deep environments \cite{liu2020underwater}. It has been used in a range of applications including monitoring of species such as corals and sponge and mapping of their habitats, identification of sediment types and cuttings from drilling operations, and identification of minerals for deep sea mining \cite{johnsen2016use}.

\begin{figure}[!t]
	\centerline{
    \includegraphics[width=0.85\linewidth]{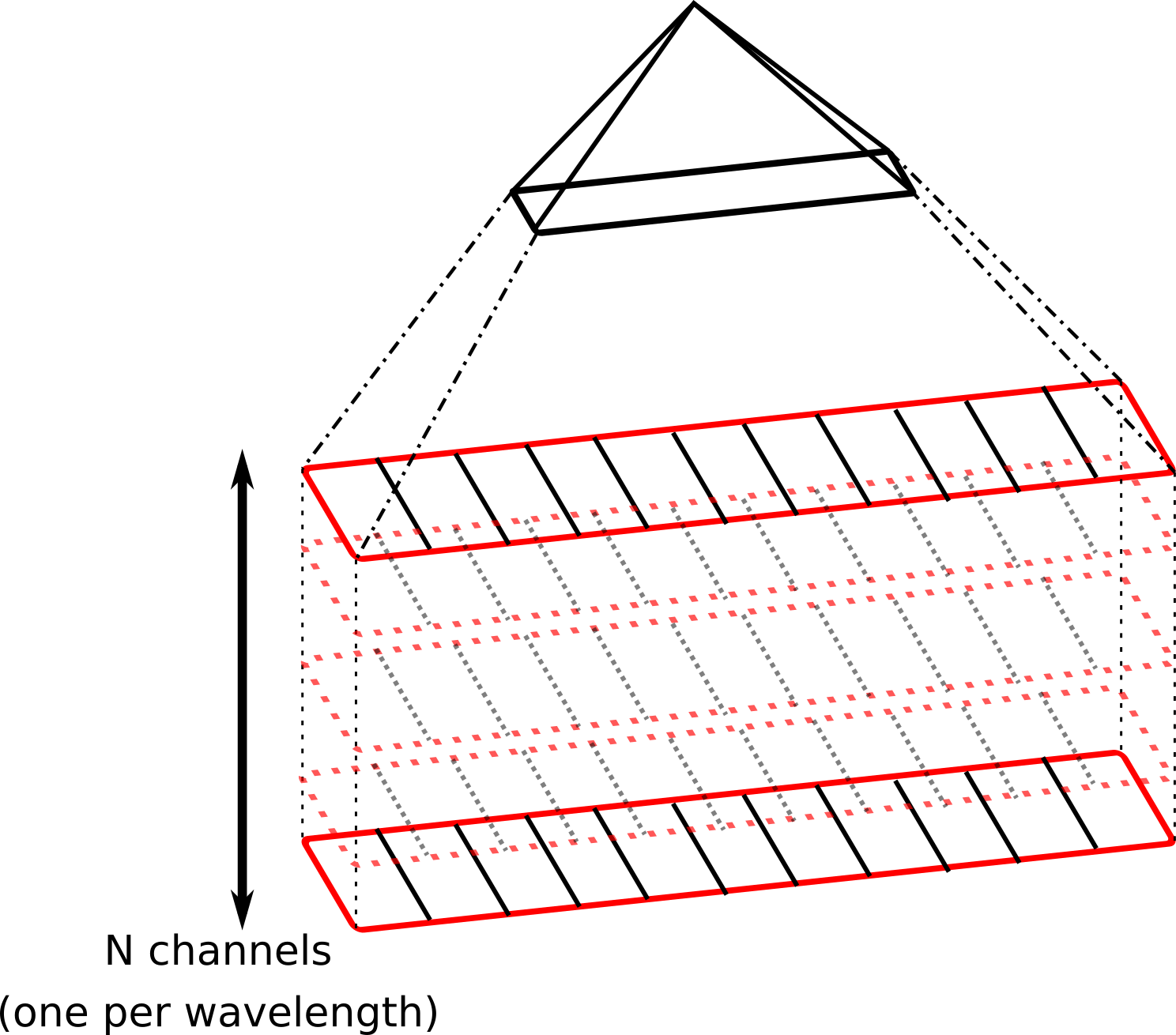}
    }
	\caption{UHI camera -- Push-broom scanner for which each image is a single array of pixels with N channels, each corresponding to a captured wavelength.}
    \label{fig:uhi_camera}
\end{figure}

In this work, we are interested in mapping the seabed using a push-broom based Underwater Hyperspectral Imaging (UHI) camera system \cite{johnsen2013uhiecotone}.  The UHI camera considered in this paper creates hyperspectral images from a single array of pixels with $N$ channels which correspond to the spectral resolution over the spectral range of the camera (see Fig.~\ref{fig:uhi_camera}).  This allows for the simulation of RGB-like images by simply extracting three channels whose wavelengths are close to the red-green-blue colors but, in practice, a much finer spectral analysis can be performed.  

The difficulty with push-broom based hyperspectral cameras comes from the fact that they capture images in a single spatial dimension.  In opposition, mosaic snapshot hyperspectral cameras capture images in two spatial directions, like classical cameras.  However, snapshot hyperspectral cameras are quite limited in terms of both spatial and spectral resolutions (typically $\leq 500 \times 500$ px and $\leq 25$ spectral bands) whereas push-broom cameras provide wide spatial and spectral resolutions (typically $\geq 960$ px and $> 100$ spectral bands) \cite{liu2020underwater}.  Despite the easier processing of mosaic snapshot hyperspectral images, push-broom based ones still provide richer and finer information.

In order to recover an exploitable material for scientific analysis, the UHI images must be processed.  The most naive way to produce understandable hyperspectral data is to sequentially stack the captured images (Fig.~\ref{fig:intro} top-left).  If one makes the assumptions of a perfectly planar scene with an ROV following a trajectory while keeping a constant depth, speed and orientation, this simple method should give acceptable results \cite{ludvigsen2014scientific}.  However, in practice none of these assumptions hold and the resulting stacked hyperspectral images exhibit strong distortions.  Hence, most works using UHI cameras have relied on mosaicking techniques to exploit and analyze the gathered hyperspectral data \cite{johnsen2013underwater,mogstad2019shallow,johnsen2016use,tegdan2015underwater,foglini2019application}.  Yet, the creation of such mosaics is performed by interpolating the UHI camera trajectory using drifting dead-reckoning navigation systems, 
leading to approximate trajectory estimation.  Furthermore, mosaicking techniques also use a flat surface assumption on the imaged scene, leading to distorted results whenever this assumption is wrong (see Fig.~\ref{fig:intro} top-right).  

This paper presents a method for creating accurate underwater hyperspectral 3D reconstructions by fusing the measurements of a UHI camera, an RGB camera and an Inertial Navigation System (INS) embedded on an ROV.  We propose to use the RGB camera to accurately estimate the trajectory followed by the ROV at the video's frame-rate 
in conjunction with the INS predictions 
for scaling purposes.  This allows us to interpolate the trajectory followed by the UHI camera. 
Additionally, we produce a dense 3D reconstruction using the acquired RGB images that we then link to the UHI images thanks to an approximately known transformation between both cameras.  Doing so, we manage to map the 3D mesh with hyperspectral textures, thus producing accurate and reliable 3D hyperspectral reconstructions that can be used to produce non-distorted ortho-images for further scientific analysis (see Fig.~\ref{fig:intro} bottom-line).

The remaining of the paper is organized as follows.  In section \ref{sec:rov_archi} we describe the architecture of the ROV employed to gather the required data.  Then, in section \ref{sec:method}, we detail the full pipeline of the proposed method.  Finally, section \ref{sec:results} presents the results obtained on the data acquired during a science cruise led by Ifremer~\footnote{\url{https://wwz.ifremer.fr/en/}}.



\section{ROV Architecture and Hyperspectral Camera}
\label{sec:rov_archi}


\begin{figure}[!t]
\begin{center}
    \includegraphics[width=\linewidth]{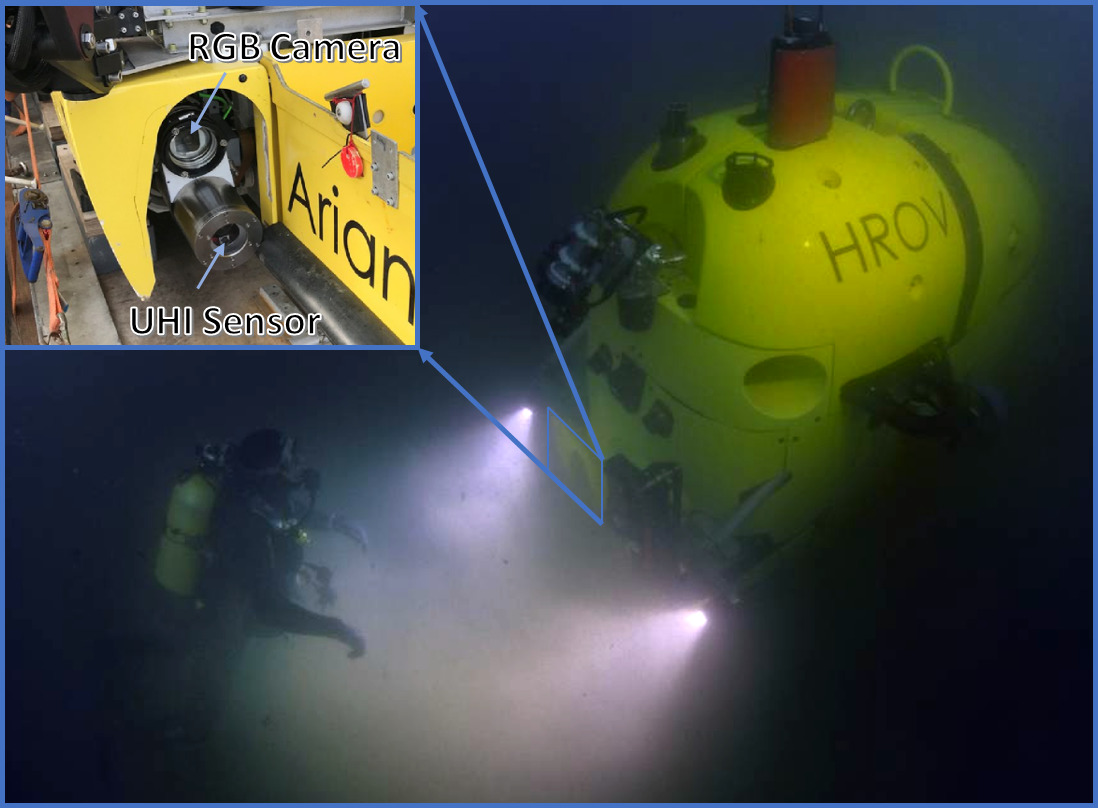}
\end{center}
\caption{This is an overview of the Ariane HROV with the UHI setup. Both the RGB video camera and the UHI where mounted on a tiltable mechanical support, at the bottom front of the vehicle.}
\vspace{-5mm}
\label{fig:hs_setup}
\end{figure}

The hyperspectral data presented in this paper has been acquired with Ifremer's Ariane HROV which can operate down to 2500m depth, in the scope of a test mission that took place in the Mediterranean sea. This is an hybrid ROV in the sense that it can both be used as an ROV or an AUV. For the ROV mode used for the hyperspectral survey, the high bandwidth link between the vehicle and the surface is provided by a reusable optical fiber. The vehicle is equipped with state of the art payloads. The optical equipment consists of multiple HD cameras and a high resolution (24Mpix) DSLR still camera. Concerning the navigation, the vehicle is equipped with high-end gyro-fiber INS, a front and a down looking Doppler Velocity Log (DVL) and an USBL positioning system. With the hybridisation of all navigation sensors, we can expect a meter accuracy localisation even in steep environment. High localisation accuracy is of prime interest for optical surveys geo-referencing and to prevent trajectory drift for the mapping algorithms.\\ 

The HROV can also be equipped with modular payloads thanks to many standard interfaces. Given this ability, we integrated an Ecotone\textsuperscript{\textregistered}~\footnote{\url{https://ecotone.com/teknologien-var-til-forskning/?lang=en}} hyperspectral camera \cite{johnsen2013uhiecotone} on the vehicle, in place of the DSLR still camera. As can be seen in Figure \ref{fig:hs_setup}, the UHI was coupled with an HD video camera and mounted on a tiltable mechanical support. This enabled us to acquire data on any type of terrain, from flat to very steep area.  Furthermore, thanks to the accurate CAD model of the HROV, we obtained an approximate 3D transformation (extrinsic calibration) between the HD and UHI cameras.

We note that, while the Ifremer's Ariane HROV has quite a specific design, the method proposed in this paper only makes use of generic underwater sensors, i.e. an RGB camera and an INS, in combination with the UHI camera.  These sensors are most common on ROVs and AUVs and the proposed pipeline is thus most likely applicable for any underwater scientific acquisition with ROVs or AUVs embedding a UHI camera.

\section{Hyperspectral 3D Mapping}
\label{sec:method}

\begin{figure*}[!t]
	\centerline{
    \includegraphics[width=\linewidth]{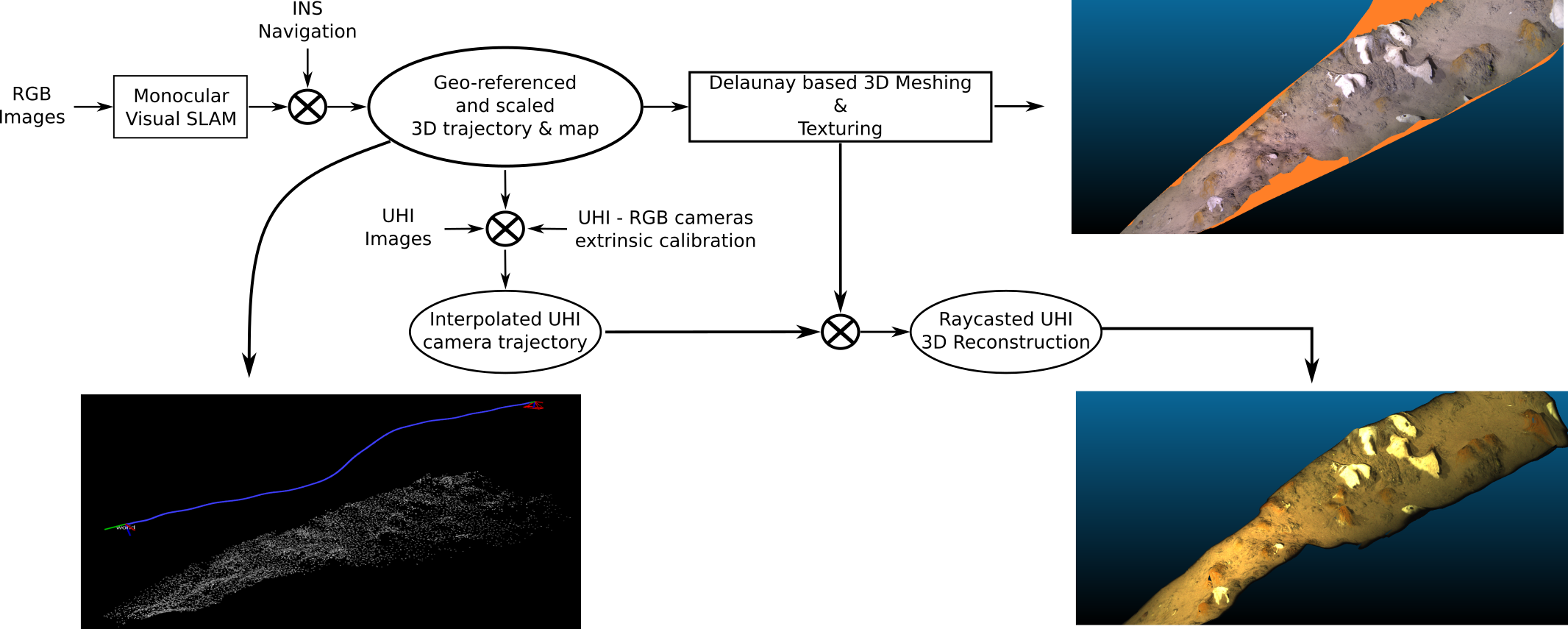}
    }
	\caption{Pipeline of the proposed method for computing hyperspectral 3D reconstructions.}
    \label{fig:pipeline}
\end{figure*}


The creation of hyperspectral 3D reconstructions requires the following information: an estimate of the trajectory followed by the UHI (i.e. an estimate of both the 3D position and orientation of the UHI for each acquired image), a 3D model of the environment that we want to map with hyperspectral textures and an estimate of the 3D transformation between the UHI and the 3D model.

We propose to leverage the RGB camera mounted on the ROV in order to estimate the trajectory followed by the UHI.  More precisely, we employ a monocular Visual Simultaneous Localization And Mapping (SLAM) algorithm to get the estimate of the trajectory followed by the RGB camera.  In addition, we fuse the visual SLAM estimates with the prediction from the INS embedded on the ROV in order to recover scaled and geo-referenced estimates.  Using the scaled SLAM results, we then compute a dense 3D mesh of the surveyed environment.  Finally, we use the approximately known extrinsic calibration between the RGB camera and the UHI in order to get an estimate of UHI trajectory and then raycast the hyperspectral images on the dense 3D model.  The proposed pipeline is illustrated in Fig.~\ref{fig:pipeline}.

\subsection{Monocular Visual SLAM for trajectory and sparse 3D map estimation}


As a push-broom scanner, the UHI must capture images at a quite high frame-rate (20 - 40 Hz) in order to sufficiently cover the scanned area. Correctly processing these images requires to know the pose of the UHI at every image.  Ideally, with a non-drifting INS 
and a ROV keeping a perfectly constant orientation and speed, the UHI poses could be estimated by means of interpolation.  However, in a real-world scenario, none of these assumptions hold.  
In order to overcome this issue, we propose to rely on the RGB camera as our main pose estimation sensor.  As the RGB camera acquire a video stream at a quite high frame-rate (25 Hz), computing the pose of this camera at every acquired image would give a very close estimation for the UHI pose.

The main solutions for estimating the trajectory followed by an RGB camera are Structure-from-Motion (SfM) and Visual SLAM (VSLAM).  On one hand, SfM is an offline technique that consists in processing a set of acquired images in order to create a 3D model by both estimating the pose of the images and a 3D reconstruction of the scene scanned.  Such techniques are at the base of photogrammetry software packages \cite{schoenberger2016sfm, moulon2016openmvg, rupnik2017micmac, arnaubec2015optical}.  Yet, these methods being computationally expensive, they are not designed to process video streams acquired at high frame-rate and do not scale well on such datasets.  On the other hand, VSLAM is a technique that is mostly used in the robotics community where having a prediction of the current robot position and orientation at almost any point in time is highly useful for numerous underlying applications \cite{mur2015orb,klein2007parallel,ferrera2019real}.  While VSLAM and SfM share similarities in the way images are processed, in VSLAM, acquired images are processed online, that is the current pose is only computed from past and current data, and a pose estimate is computed for every image.  While VSLAM might be slightly less accurate than SfM because of its limitation to past and present measurements at any given point (whereas in SfM all the measurements are processed simultaneously), it is actually a very reliable solution in this context as it copes with the camera's video frame-rate and thus provides an accurate basis for the computation of the UHI images' pose.

In this work, we have used the open-source algorithm \ov2slam~\footnote{\url{https://github.com/ov2slam/ov2slam}} \cite{ferrera2021ov} to perform VSLAM on the video sequences.  \ov2slam uses optical flow \cite{baker2004lucas} to track keypoints over the image sequences and use these tracks to both estimate the camera's poses and a 3D map of the imaged scene.  Note that we have used \ov2slam here because we found out that it performs well on underwater data but in theory any other VSLAM method could be used.  We refer the interested readers to \cite{ferrera2021ov} for more information on the algorithm.  An example of the trajectory and 3D map that we estimate using the VSLAM algorithm is shown on the bottom-left of Fig.~\ref{fig:pipeline}. 

Nonetheless, in a monocular configuration (i.e. using only one RGB camera), the trajectories and 3D maps estimated by SfM and VSLAM methods are limited by a scale ambiguity and hence the estimations are performed up to an unknown scale factor which must be recovered to convert these estimations on a metric scale.  Furthermore, it is often useful to have geo-referenced estimations for further analysis.  In the following section we detail how we manage to both scale and geo-reference the VSLAM estimations thanks to the INS.

\subsection{Scaling and geo-referencing by fusing INS predictions}


The INS geo-referenced prediction of the ROV position has been extracted at a rate of 1 Hz for this survey. These predictions are given in the geodetic latitude-longitude-altitude format but they can be easily converted to produce relative positions of the ROV in a North-East-Down world frame.

Using the VSLAM algorithm, we obtain a pose estimate $\mathbf{T_{wc_{i}}} \in \SE3$ for each processed image, where $\SE3$ denotes the 3D Special Euclidean group \cite{barfoot2017state}, a set of 3D map points $\boldsymbol{\lambda}^{w}_{k} \in \mathbb{R}^{3}$ and a set of 3D map points 2D observations per image $\mathbf{x_{ik}} \in \mathbb{R}^{2}$.
Considering a calibrated camera, these state parameters are related by the projection function $\boldsymbol{\pi} : \mathbb{R}^{3} \mapsto \mathbb{R}^{2}$:

\begin{gather}
    \mathbf{x}_{ik} = \boldsymbol{\pi} \left(\mathbf{T_{c_{i}w}} \odot \boldsymbol{\lambda}^{w}_{k}  \right) \\
    \mathbf{x}_{ik} = \boldsymbol{\pi} \left(\mathbf{R_{c_{i}w}} \cdot \boldsymbol{\lambda}^{w}_{k} + \mathbf{t_{c_{i}w}} \right)
\end{gather}

where $\mathbf{T_{c_{i}w}} = \mathbf{T^{-1}_{wc_{i}}}$ and $\mathbf{R_{c_{i}w}}$ and $\mathbf{t_{c_{i}w}}$ are respectively the rotation matrix and the translation part of image $i$ inverse pose $\mathbf{T_{c_{i}w}}$. 

We can relate a subset of images to the INS predictions by taking the images closest in time to a given prediction.  From those, we can define a nonlinear optimization problem where we seek to align the images' poses and the 3D map points with the geo-referenced predictions of the INS.  We denote the full set of images' poses linked to an INS prediction $\mathbf{p_i} \in \mathbb{R}^{3}$ as $\boldsymbol{\zeta}$.  We further include in $\boldsymbol{\zeta}$ all the 3D map points observed by the images already in $\boldsymbol{\zeta}$.  We can now define a Bundle Adjustment like nonlinear optimization problem made of a collection of 2D reprojection errors along with a collection of 3D positioning errors:


\begin{gather*}
    \boldsymbol{\zeta}^{*} = \argmin_{\boldsymbol{\zeta}}  \sum_{i \in \mathfrak{K}} \sum_{k \in \mathfrak{L}_i} \left \| \mathbf{x}_{ik} - \boldsymbol{\pi} \left( \mathbf{T_{c_{i}w}} \odot \boldsymbol{\lambda}^{w}_{k} \right)  \right \|^{2}_{\boldsymbol{\Sigma}}  \\
    + \sum_{i \in \mathfrak{K}} \left \| \mathbf{p}_{i} - \mathbf{t_{wc_{i}}} \right \|^{2}_{\boldsymbol{\Sigma'}} 
\end{gather*}

where $\mathfrak{K}$ is the set of keyframes in $\boldsymbol{\zeta}$ and $\mathfrak{L}_i$ is the set of 3D map points observed by keyframe $i$.  The covariances $\boldsymbol{\Sigma}$ and $\boldsymbol{\Sigma'}$ are respectively associated to the 2D visual observations and to the INS predictions.

In practice we solve this optimization problem using the Levenberg-Marquardt algorithm \cite{kanzow2005withdrawn}.  Once the optimization has converged, we obtain a fully scaled and geo-referenced 3D maps.  However, we only have geo-referenced pose estimates for the subset of the images included in $\boldsymbol{\zeta}$.  While we could have added all the images to the previous optimization problem, it would have lead to a huge problem difficult to solve on standard computers.  Instead, we simply compute the new pose of every remaining images using their known 2D observations of now geo-referenced 3D map points by applying a the Perspective-from-N-Points method \cite{hartley2003multiple}.  Note that, if geo-referencing is not required, any other sensor able to provide a scaling information could be used in place of an INS here \cite{istenic2019scale,istenic2020automatic,ferrera2019aqualoc}.

We can now use the extrinsic calibration to interpolate the geo-referenced trajectory of the UHI from the RGB camera estimated trajectory.  Assuming a flat surface, this result would be sufficient to produce hyperspectral photo-mosaics.  However, as we want to create 3D reconstruction that strictly respects the reality of the surveyed environments, we still need to produce a dense 3D model from the sparse 3D map estimated by the VSLAM and a way for texturing this 3D model with hyperspectral intensities.

\subsection{Hyperspectral 3D reconstruction}


In order to extract a dense 3D mesh from the sparse 3D map computed in the previous stage, we apply a 3D Delaunay tetrahedralization followed by a graph-cut method \cite{jancosek2014exploiting} using the open-source OpenMVS library~\footnote{\url{https://github.com/cdcseacave/openMVS}} \cite{openmvs2020}.  The dense 3D mesh obtained in this step is then textured using the method of \cite{waechter2014let}.  The top-right of Fig.~\ref{fig:pipeline} displays an example of the textured 3D mesh we manage to recover.  

In order to now apply hyperspectral textures to the mesh, we propose to raycast the hyperspectral images on the 3D mesh.  To do so, we project a ray for every pixel in every hyperspectral image onto the 3D mesh in order to find the coordinates of its intersection with the mesh and hence get a depth value for each pixel.  Henceforth, after this step we obtain the equivalent of a depth map for all the hyperspectral images that we can directly use along with the known UHI poses to produce a dense 3D pointcloud with hyperspectral intensities (bottom-right of Fig.~\ref{fig:pipeline}).

\begin{figure}[!t]
    \centering
    \begin{subfigure}[b]{0.49\linewidth}
        \centering \includegraphics[width=\linewidth]{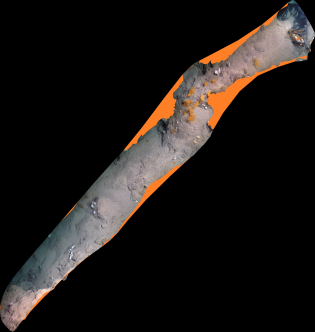}
        \captionsetup{justification=centering}
        \caption{RGB ortho-image.}\label{fig:ortho_rgb}
    \end{subfigure}
    \hfill
    \begin{subfigure}[b]{0.49\linewidth}
        \centering \includegraphics[width=\linewidth]{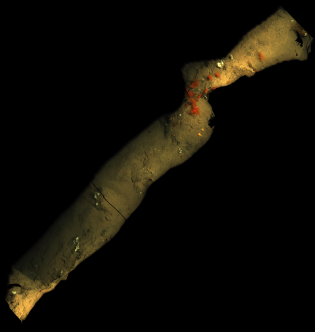}
        \captionsetup{justification=centering}
        \caption{Hyperspectral ortho-image.}\label{fig:ortho_hs}
    \end{subfigure}
    
    \begin{subfigure}[b]{\linewidth}
        \centering \includegraphics[width=\linewidth]{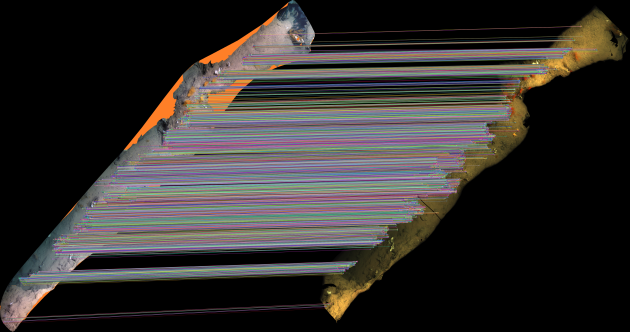}
        \captionsetup{justification=centering}
        \caption{Features matched between the RGB and the hyperspectral ortho-images.}\label{fig:ortho_klt}
    \end{subfigure}
    
    \captionsetup{justification=centering}
    \caption[Sample results top-view.]{Features matching between RGB and hyperspectral ortho-images for extrinsic calibration refinement.}
    \label{fig:ortho_klt_rgb_hs}
\end{figure}

While the proposed method already produces high quality 3D maps, the uncertainty of the extrinsic calibration in addition to the measurements' noise contributed to some mis-alignment between the RGB  and the hyperspectral textures. The extrinsic calibration $\mathbf{T_{rel}}$ between the RGB and UHI cameras is thus additionally refined. By leveraging on the dense 3D reconstructions obtained in the previous stages and converting them into 2D ortho-images, associations between some pixels from the initial UHI images and known 3D map points in the frame of reference can be obtained through feature matching between both ortho-images as shown in Fig.~\ref{fig:ortho_klt_rgb_hs}. In our case we have used ORB features \cite{rublee2011orb}.  By relating known 3D coordinates and UHI ortho-image's pixels, the extrinsic calibration can be refined through nonlinear optimization such as Bundle Adjustment in which the mis-alignment is minimized.

In such approach, the accuracy of such RGB-UHI correspondences depend on the resolution of the produced ortho-images. In our case, we have created the ortho-images with a resolution of 5 $\times$ 5 mm per pixel which led to satisfying results.


\section{Experimental Results}
\label{sec:results}

The results presented in this section have been acquired with the HROV Ariane (Fig.~\ref{fig:hs_setup}) while performing close to seabed transects.  We have used a UHI camera from Ecotone\textsuperscript{\textregistered} \cite{johnsen2013uhiecotone} running at 30 Hz and capturing $1920 \times 1$ px images with 211 spectral bands, from which 3 bands have been extracted to simulate red-green-blue colors in the following results.  The HD camera captured RGB images at 25 Hz which have been down-sampled into $960 \times 540$ px images before processing and the INS output predictions at 1 Hz.

All the developments have been made in C++ and rely on open-source software and libraries: \ov2slam \cite{ferrera2021ov} for the VSLAM, OpenMVS \cite{openmvs2020} for the RGB 3D mesh reconstruction, OpenCV \cite{opencv} for image processing tasks and Ceres \cite{ceres-solver} for nonlinear optimization operations.


\begin{figure}[!t]
\begin{center}
    \includegraphics[width=\linewidth]{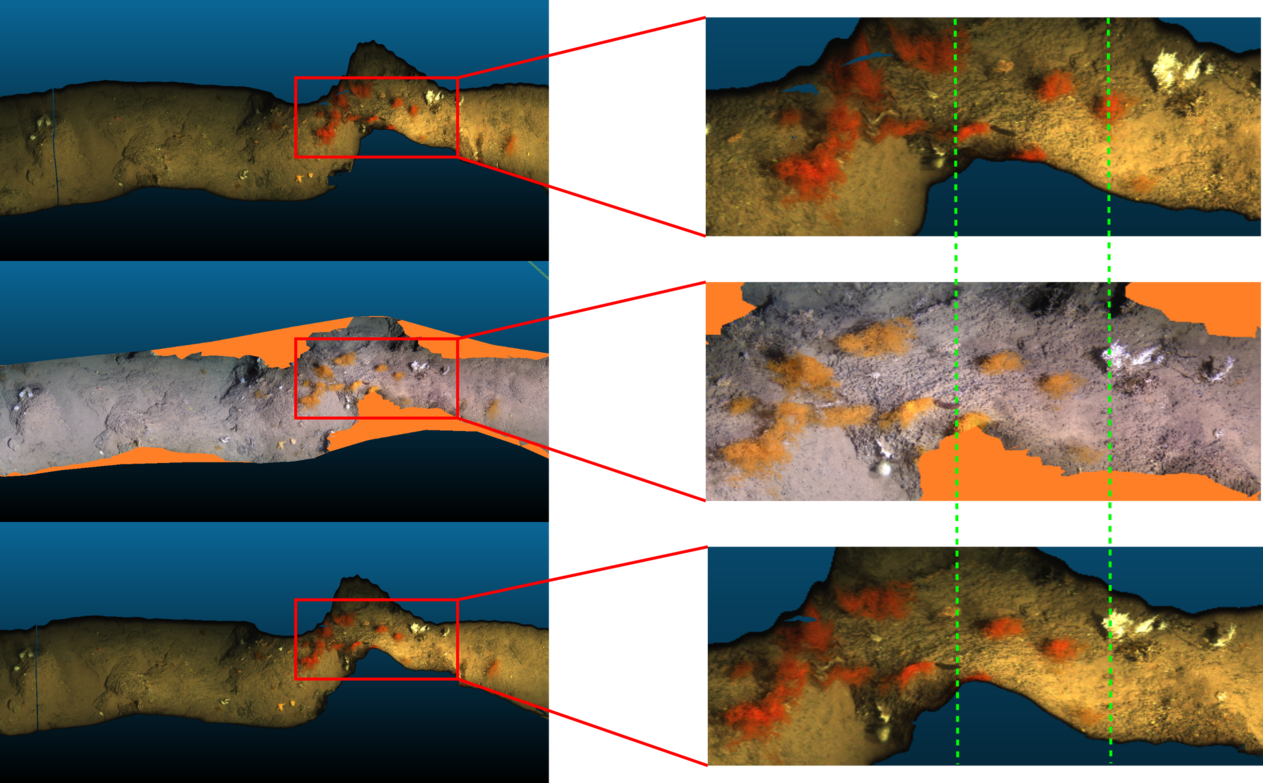}
\end{center}
\caption{Hyperspectral - RGB refinement.  From top to bottom: initial hyperspectral mesh, RGB mesh, refined hyperspectral mesh.}
\vspace{-5mm}
\label{fig:hs_rgb_refinement}
\end{figure}

\subsection{Qualitative results}

To assess the efficiency of the method we propose for creating hyperspectral 3D reconstructions of underwater environments, we show results obtained on four different datasets.  For every dataset we compare the hyperspectral 3D model that we reconstruct to both the RGB 3D model obtained on the same sequence and to the initial UHI images simply stacked horizontally. The benefit of applying the extrinsic calibration refinement step is shown in Fig.~\ref{fig:hs_rgb_refinement}, where the improved result is shown together with the initial mis-alignement between the RGB and hyperspectral reconstructions.

\begin{figure}
    \centering

    \begin{subfigure}[b]{\linewidth}
        \centering \includegraphics[width=\linewidth]{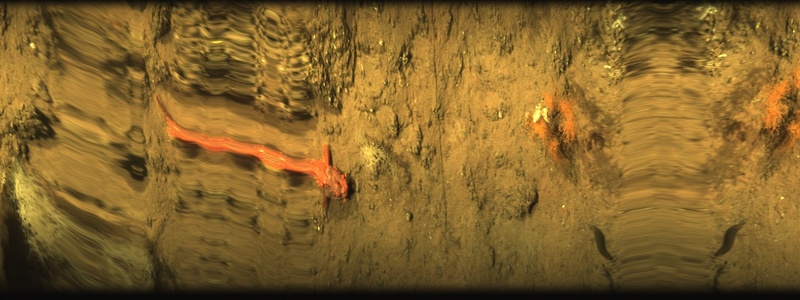}
        \captionsetup{justification=centering}
        \caption{Initial stacked UHI images.}\label{fig:hs_img_2}
    \end{subfigure}
    ~
    \begin{subfigure}[b]{\linewidth}
        \centering \includegraphics[width=\linewidth]{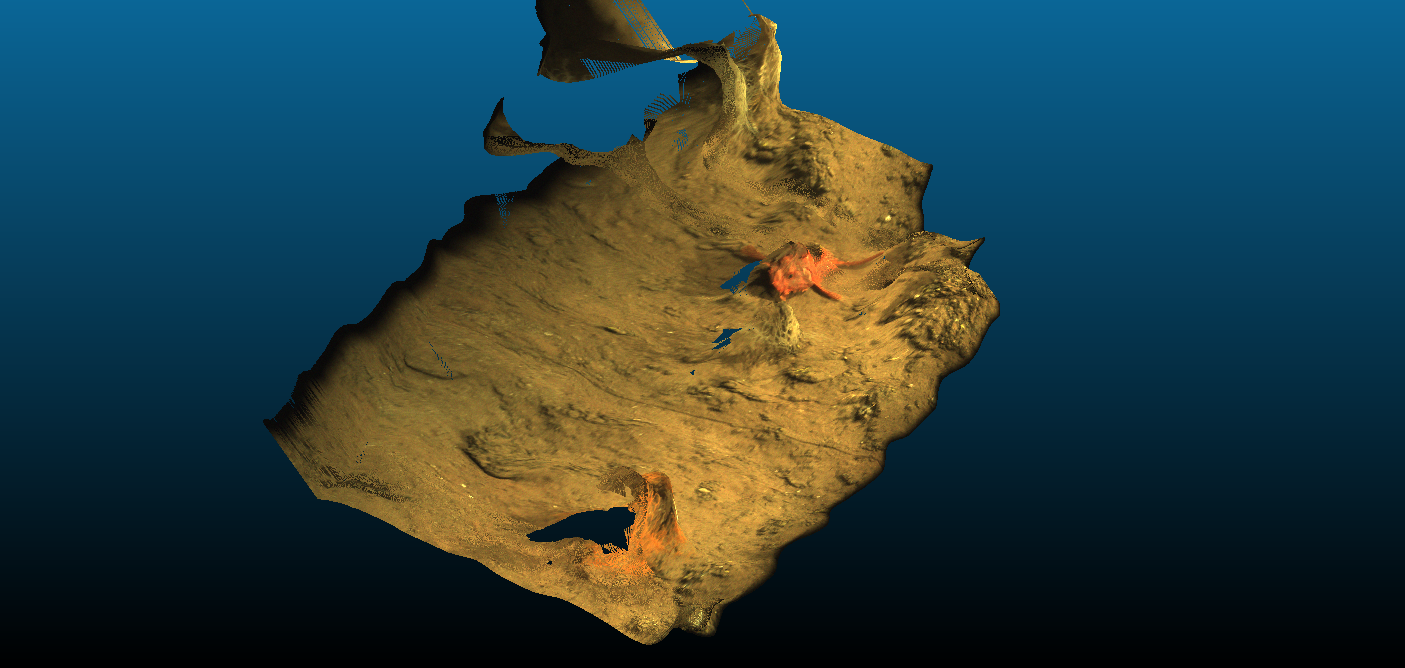}
        \captionsetup{justification=centering}
        \caption{Reconstructed Hyperspectral 3D Mesh.}\label{fig:hs_mesh_2}
    \end{subfigure}
    ~
    \begin{subfigure}[b]{\linewidth}
        \centering \includegraphics[width=\linewidth]{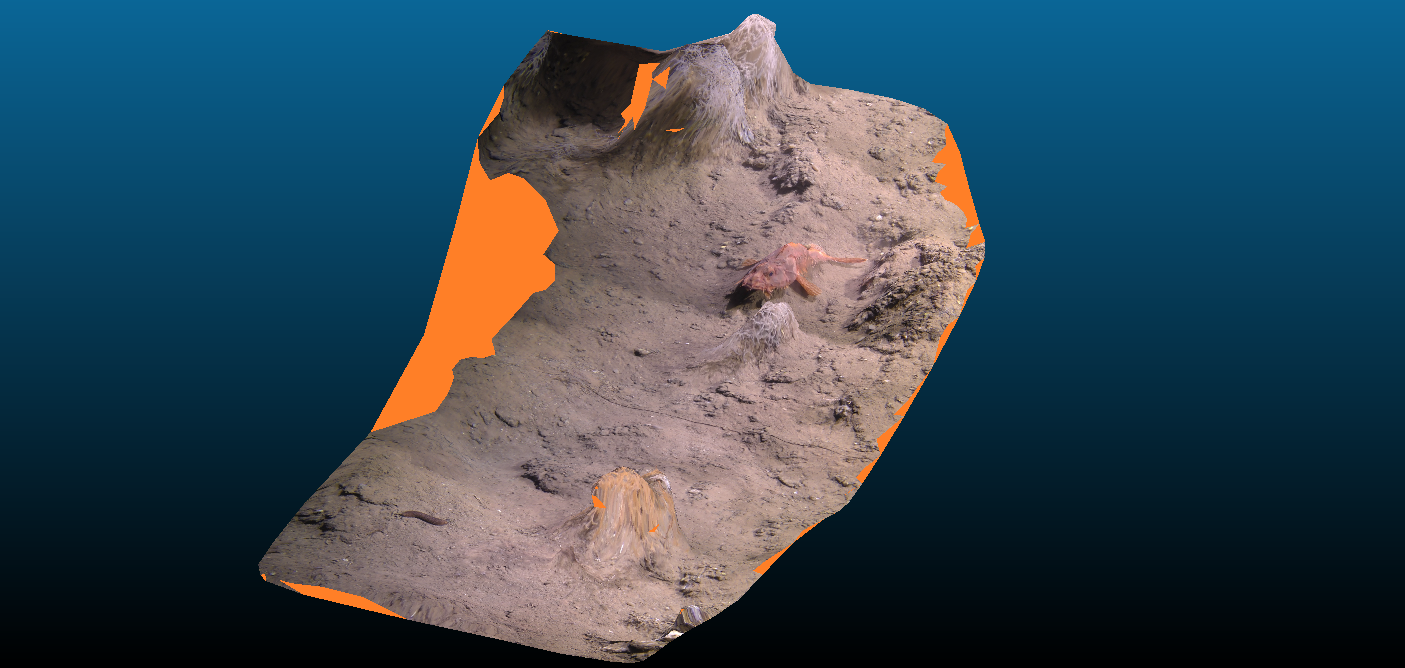}
        \captionsetup{justification=centering}
        \caption{Reconstructed RGB 3D Mesh.}\label{fig:rgb_mesh_2}
    \end{subfigure}
    \captionsetup{justification=centering}
    \caption{Comparison of the final hyperspectral and RGB 3D reconstructions obtained on a small scene.}
    \label{fig:rgb_hs_mesh_2}
\end{figure}

The first dataset is a small scene that highlights well the benefit of our method for producing reliable 3D hyperspectral reconstructions.  The results obtained on this dataset are illustrated in Fig.~\ref{fig:rgb_hs_mesh_2}.  It can be seen that the distortion effects that are visible on the stacked UHI images are well removed in the final 3D reconstruction.

\begin{figure}[!t]
    \centering
     \includegraphics[width=\linewidth]{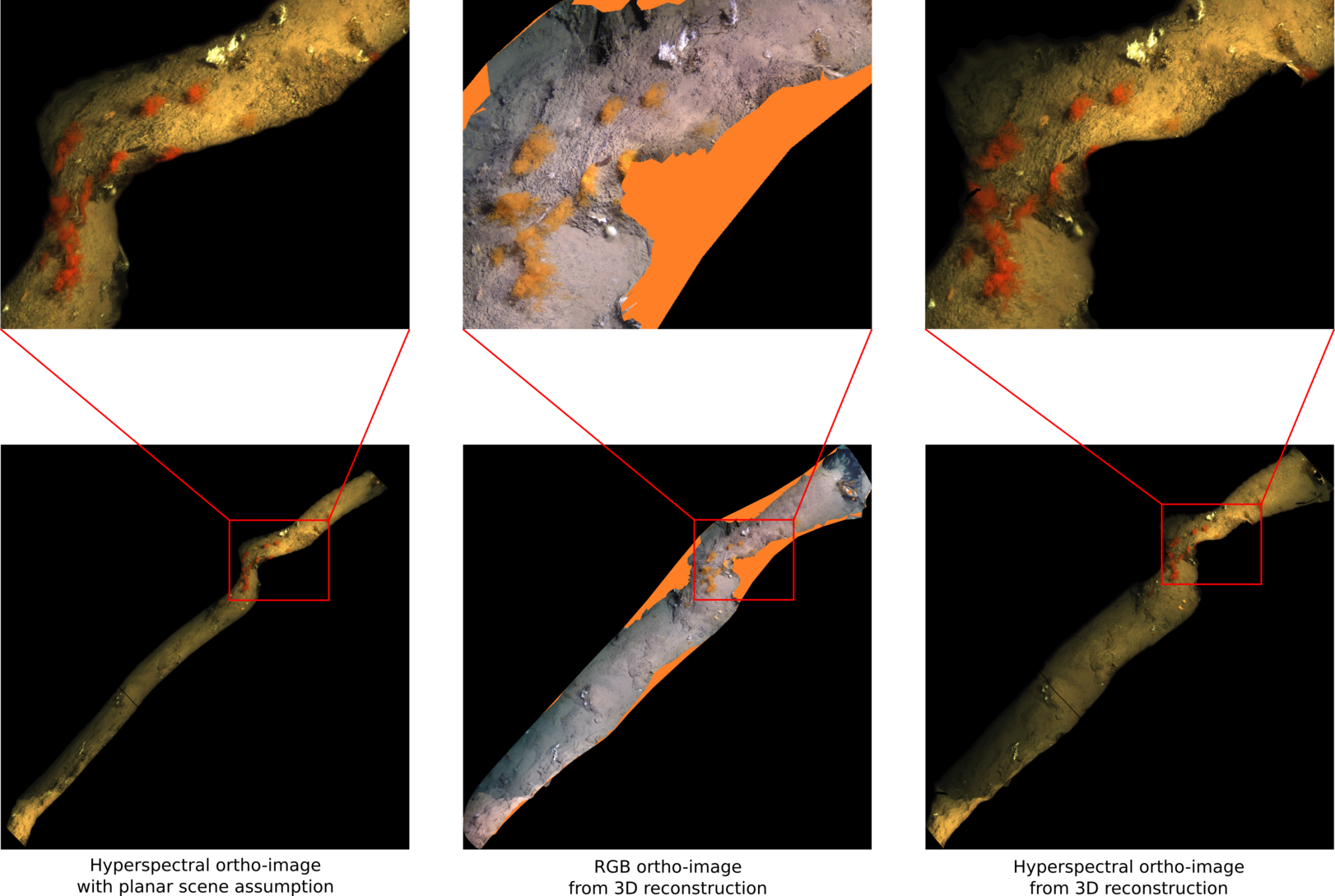}
    \caption{Ortho-image from planar assumption versus ortho-image from 3D reconstruction.}
    \label{fig:ortho_planar}
\end{figure}

\begin{figure}
    \centering

    \begin{subfigure}[b]{\linewidth}
        \centering \includegraphics[width=\linewidth]{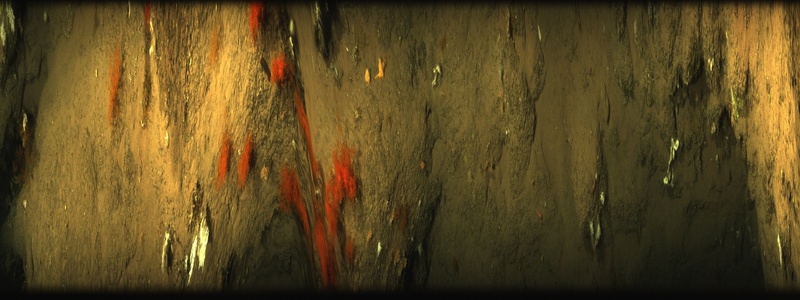}
        \captionsetup{justification=centering}
        \caption{Initial stacked UHI images.}\label{fig:hs_img_1}
    \end{subfigure}

    \begin{subfigure}[b]{\linewidth}
        \centering \includegraphics[width=\linewidth]{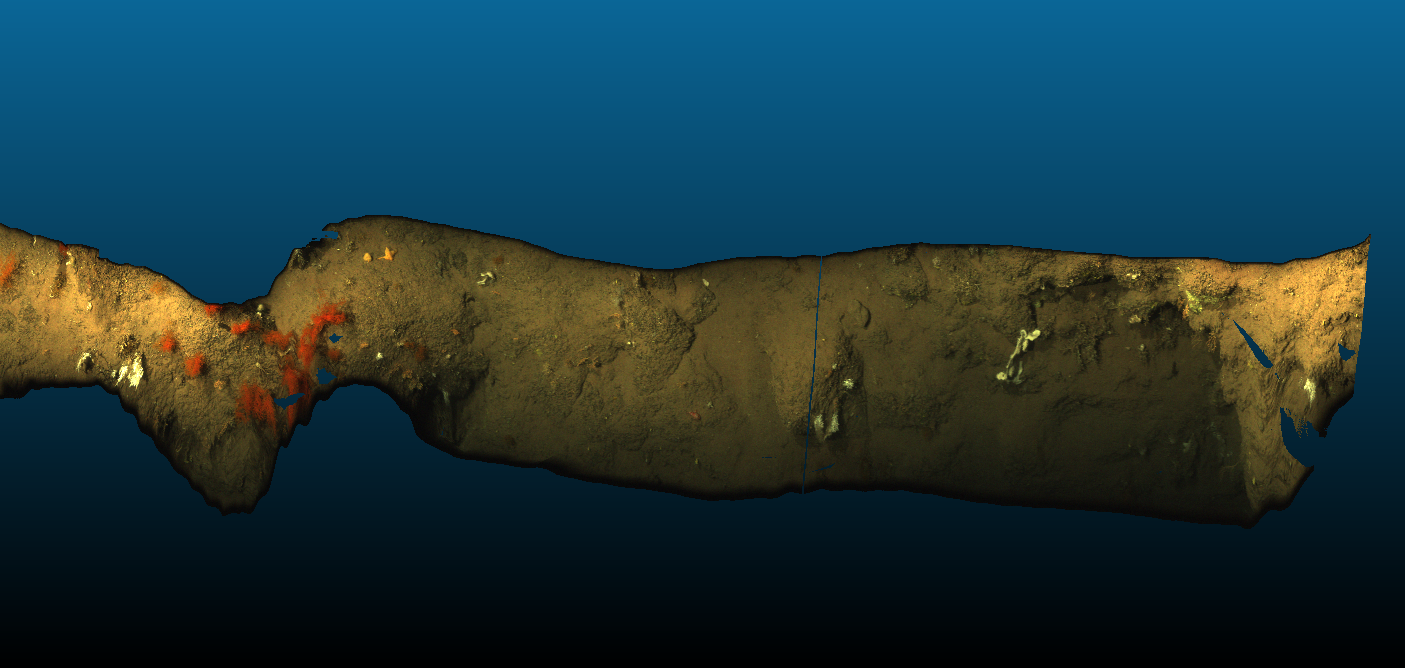}
        \captionsetup{justification=centering}
        \caption{Hyperspectral 3D Mesh (top-view).}\label{fig:hs_mesh_1_bis}
    \end{subfigure}

    
    \begin{subfigure}[b]{\linewidth}
        \centering \includegraphics[width=\linewidth]{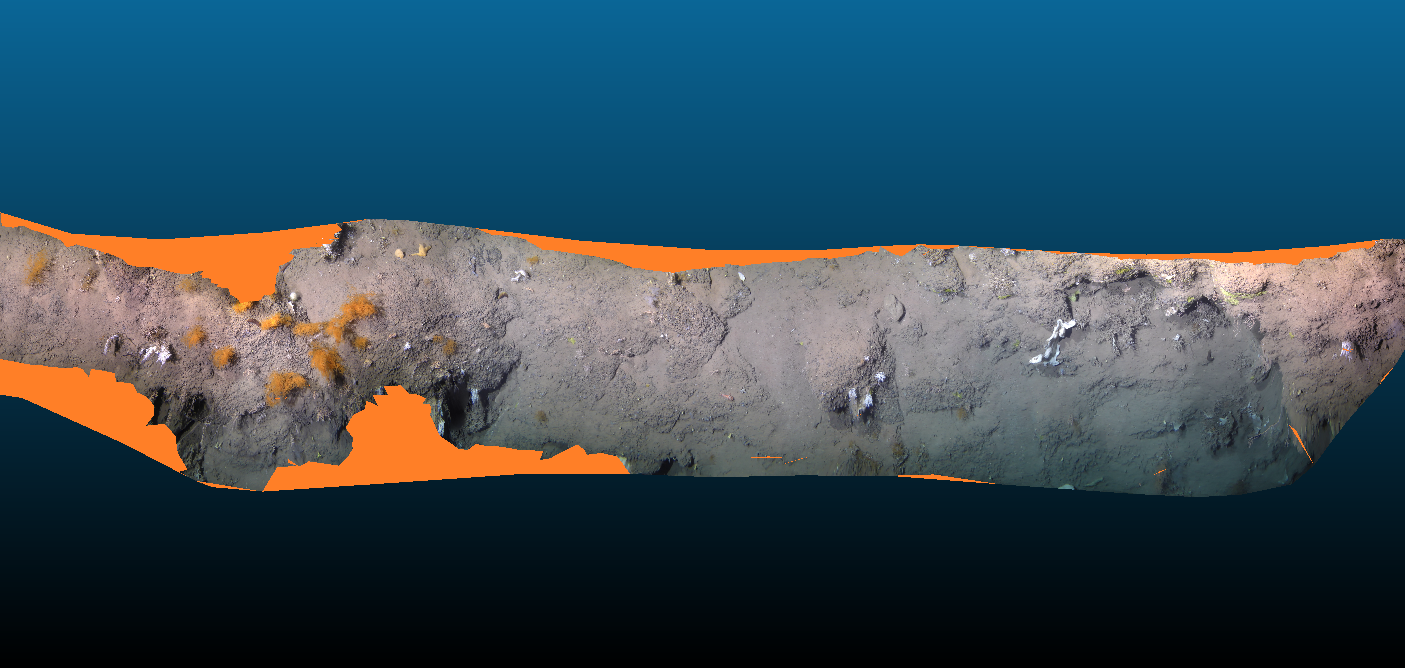}
        \captionsetup{justification=centering}
        \caption{RGB 3D Mesh (top-view).}\label{fig:rgb_mesh_1_bis}
    \end{subfigure}
    
    \begin{subfigure}[b]{\linewidth}
        \centering \includegraphics[width=\linewidth]{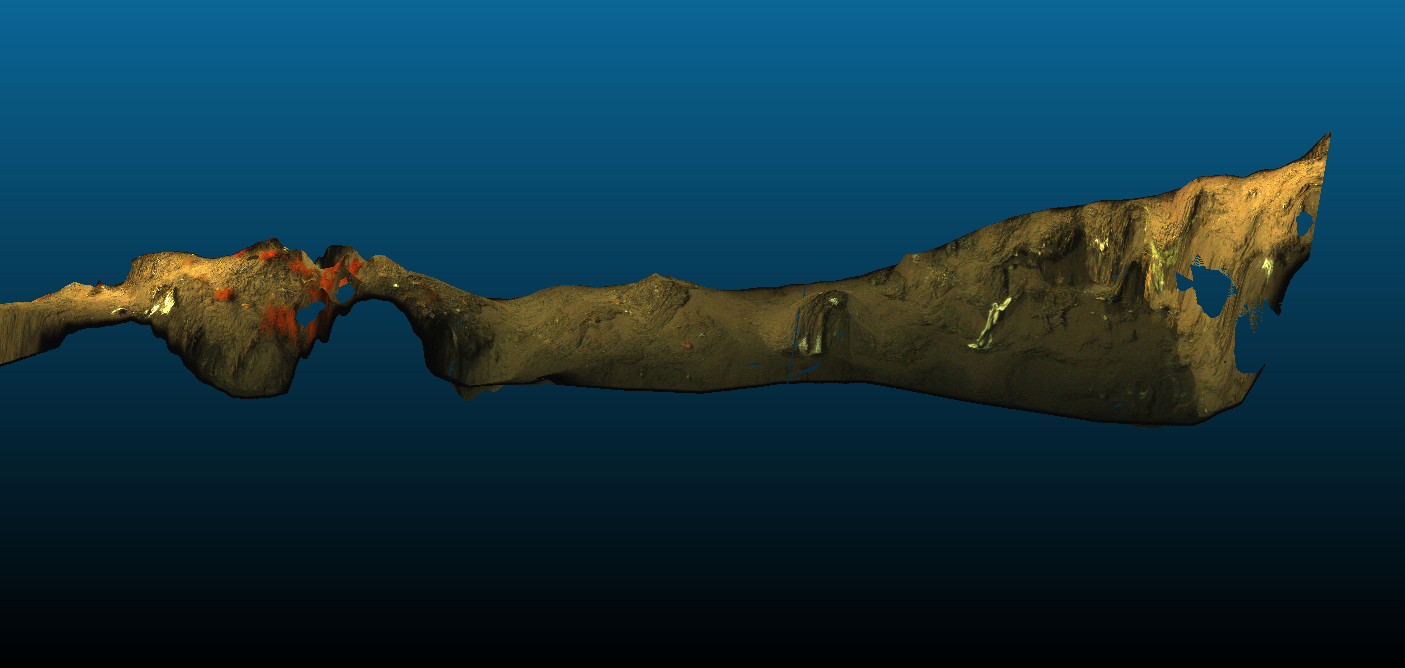}
        \captionsetup{justification=centering}
        \caption{Hyperspectral 3D Mesh (side-view).}\label{fig:hs_mesh_1}
    \end{subfigure}

    \begin{subfigure}[b]{\linewidth}
        \centering \includegraphics[width=\linewidth]{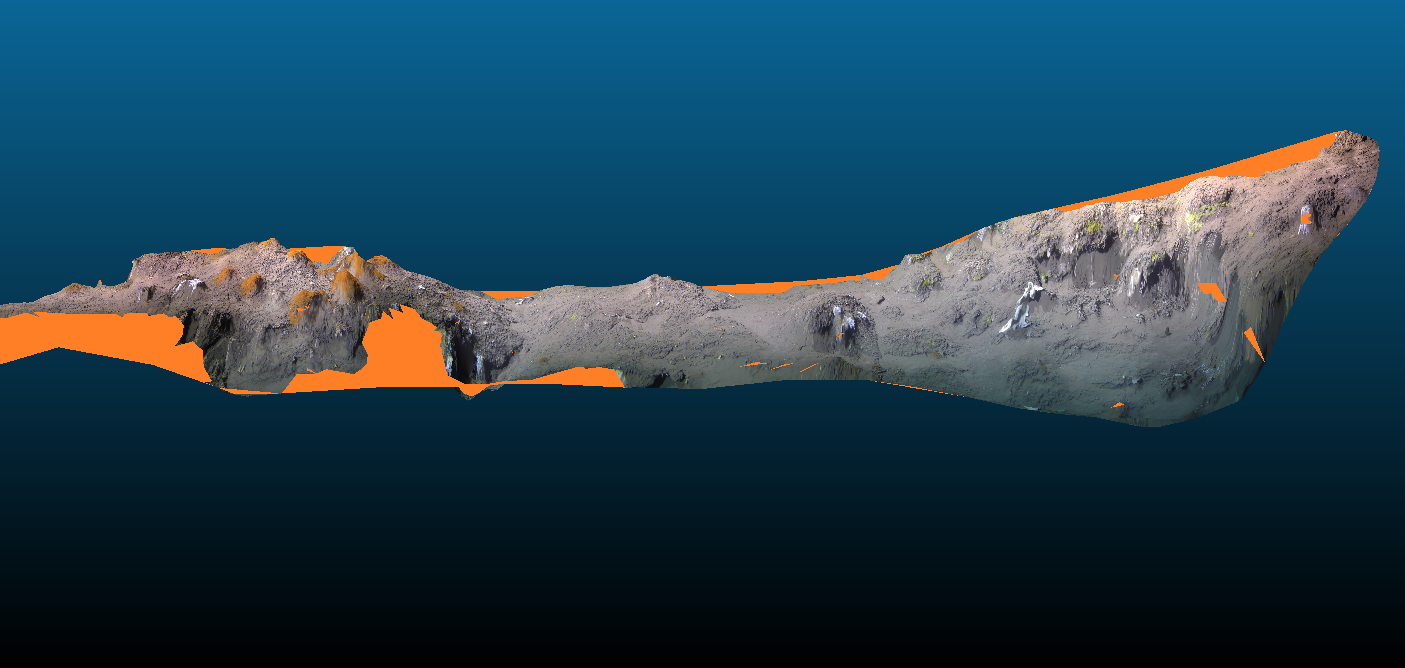}
        \captionsetup{justification=centering}
        \caption{RGB  3D Mesh (side-view).}\label{fig:rgb_mesh_1}
    \end{subfigure}



    \captionsetup{justification=centering}
    \caption{Comparison of the final hyperspectral and RGB 3D reconstructions obtained on a scene with low 3D.}
    \label{fig:rgb_hs_mesh_1}
\end{figure}

The second dataset comes from a transect performed while hovering above a scene with low but non-negligeable 3D.  Fig.~\ref{fig:rgb_hs_mesh_1} shows the results obtained on this dataset.  Once again, the results have a high fidelity with the 3D RGB reconstruction.  Fig.~\ref{fig:ortho_planar} highlights the fact that even on such scene with low 3D variations, discarding the planar assumption leads to significantly better results, with almost no distortion compared to methods based on a planar scene assumption.

\begin{figure}
    \centering

    \begin{subfigure}[b]{\linewidth}
        \centering \includegraphics[width=\linewidth]{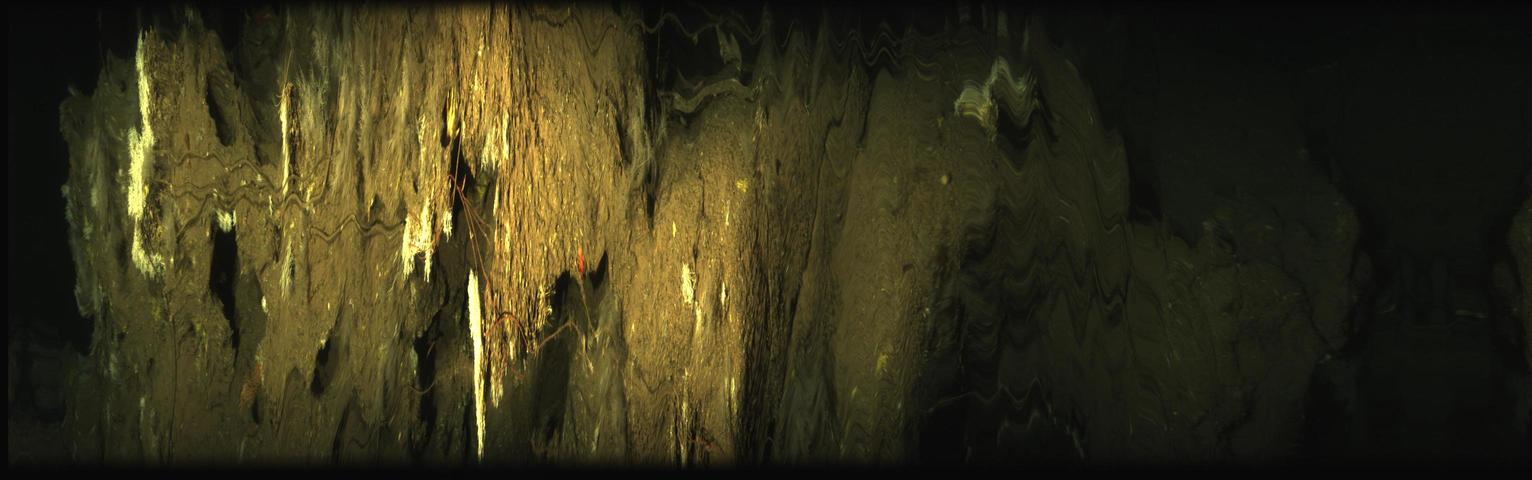}
        \captionsetup{justification=centering}
        \caption{Initial stacked UHI images.}\label{fig:hs_img_3}
    \end{subfigure}

    \begin{subfigure}[b]{\linewidth}
        \centering \includegraphics[width=\linewidth]{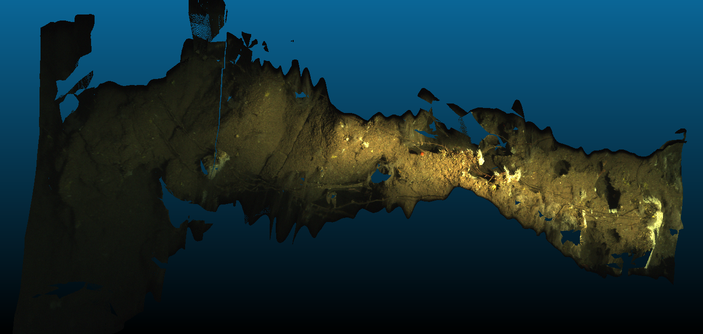}
        \captionsetup{justification=centering}
        \caption{Hyperspectral 3D Mesh (top-view).}\label{fig:hs_mesh_3_bis}
    \end{subfigure}
    
    \begin{subfigure}[b]{\linewidth}
        \centering \includegraphics[width=\linewidth]{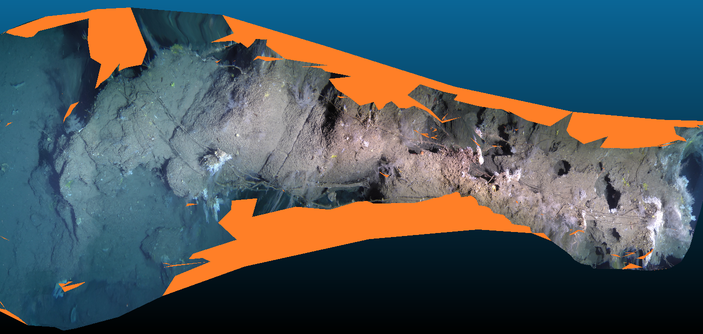}
        \captionsetup{justification=centering}
        \caption{RGB 3D Mesh (top-view).}\label{fig:rgb_mesh_3_bis}
    \end{subfigure}
    
    \begin{subfigure}[b]{\linewidth}
        \centering \includegraphics[width=\linewidth]{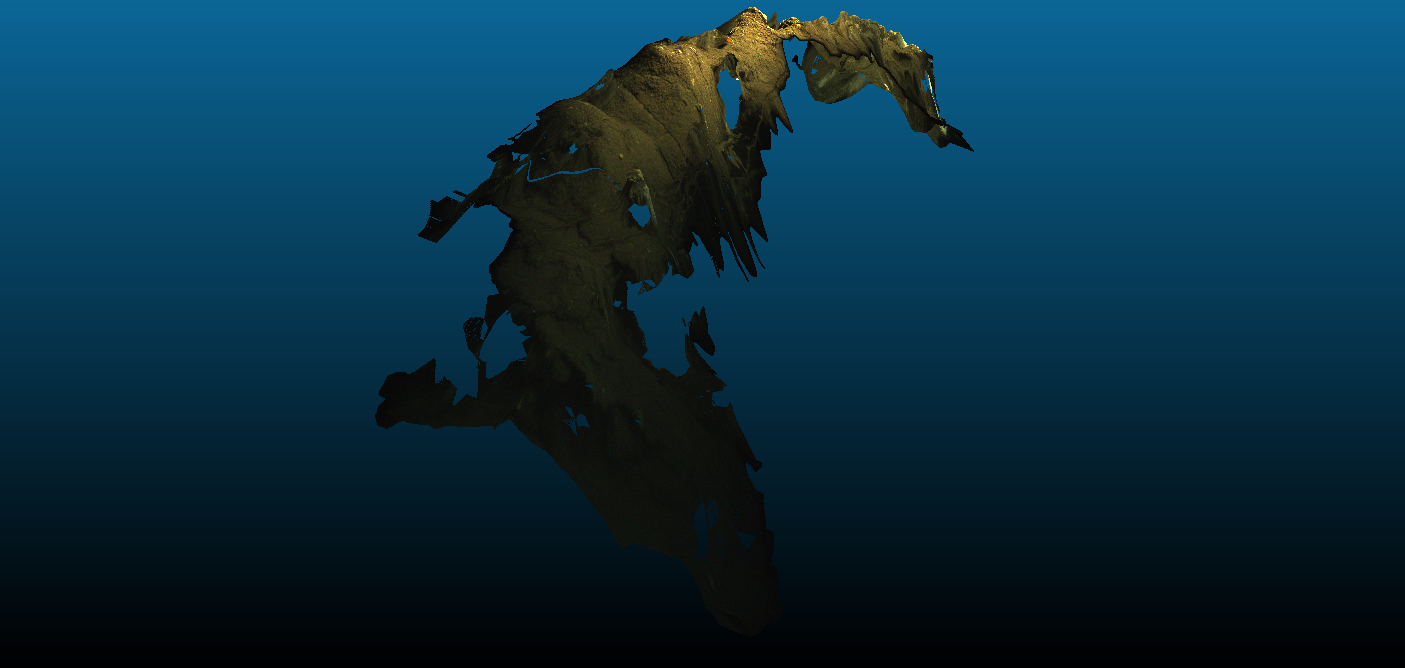}
        \captionsetup{justification=centering}
        \caption{Hyperspectral 3D Mesh (side-view).}\label{fig:hs_mesh_3}
    \end{subfigure}

    \begin{subfigure}[b]{\linewidth}
        \centering \includegraphics[width=\linewidth]{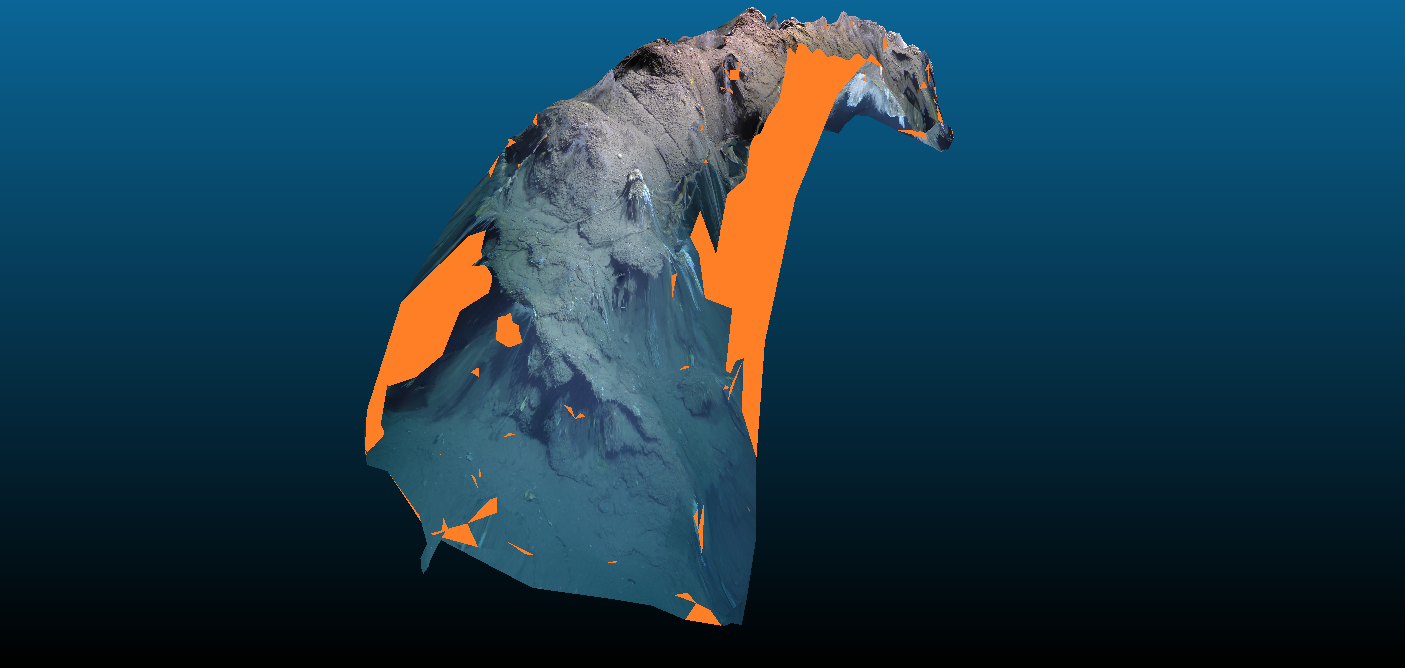}
        \captionsetup{justification=centering}
        \caption{RGB 3D Mesh (side-view).}\label{fig:rgb_mesh_3}
    \end{subfigure}



    \captionsetup{justification=centering}
    \caption{Comparison of the final hyperspectral and RGB 3D reconstructions obtained on a scene with significant 3D variations of its structure.}
    \label{fig:rgb_hs_mesh_3}
\end{figure}

The third dataset is a scene with strong 3D variations.  Fig.~\ref{fig:rgb_hs_mesh_3} displays the results obtained on this dataset.  Comparing to the ortho-image obtained with a planar scene assumption in Fig.~\ref{fig:intro}, the obtained hyperspectral reconstruction is clearly less distorted and would make a more effective material to work on for scientists.

\begin{figure}
    \centering
     \includegraphics[width=\linewidth]{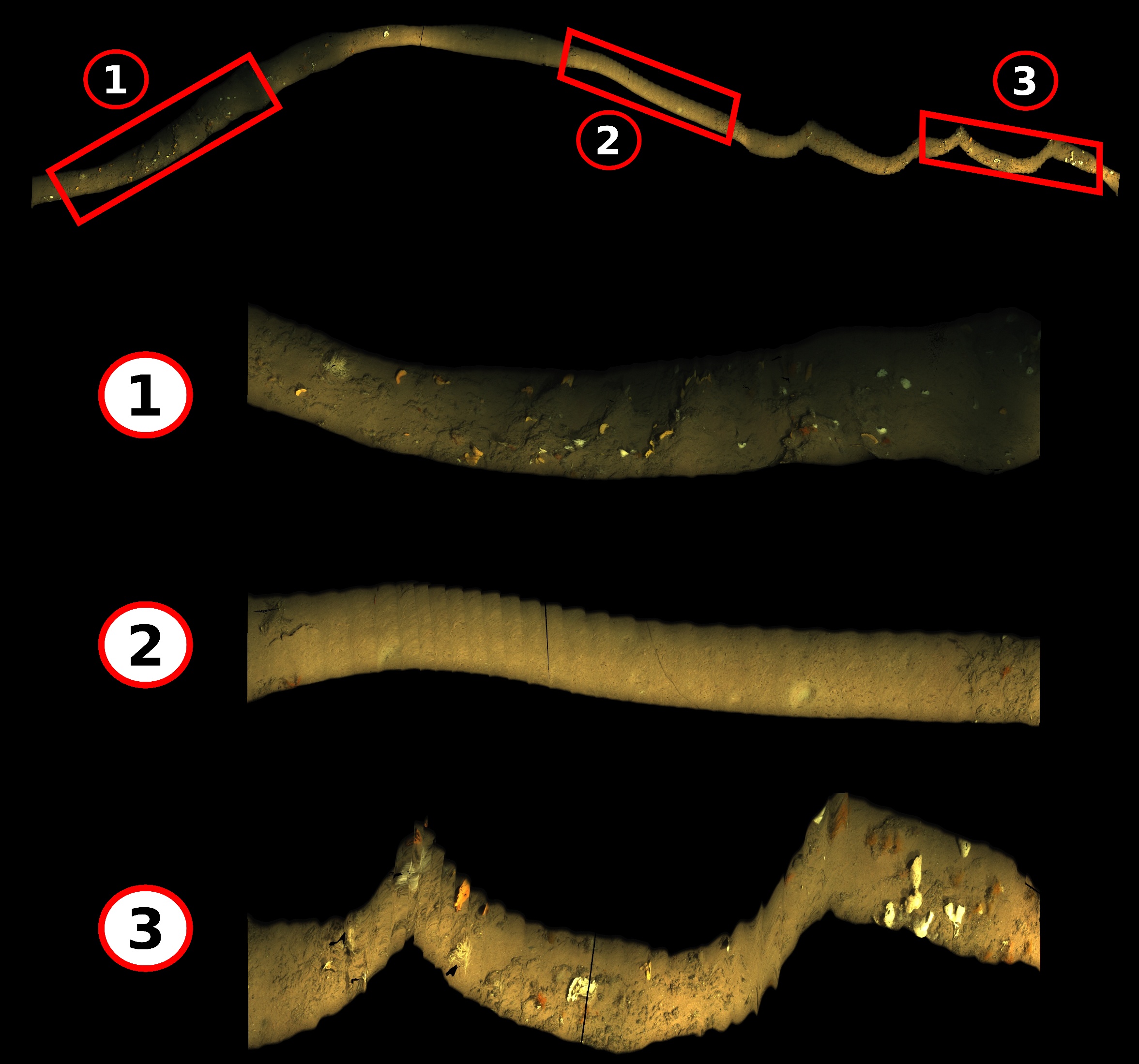}
    \caption{Hyperspectral ortho-image on a more than 100 meters long transect.}
    \label{fig:scalability}
\end{figure}

The last dataset consists of a large transect more than 100 meters long to highlight the scalability of our method.  The resulting hyperspectral ortho-image computed from the 3D reconstruction is shown in Fig.~\ref{fig:scalability}.



\section{Discussion}

The underwater hyperspectral 3D mapping method we propose in this paper is, to the best of our knowledge, the first method of this kind.  Taking advantage over the great advances in 3D reconstructions from RGB images in the computer vision and robotics communities, we manage to create accurate and reliable hyperspectral 3D reconstructions. While we have successfully addressed the issue of extrinsic RGB-UHI calibration, additional improvements such as avoiding matching through 2D mosaics can be made to increase the robustness and accuracy. Another remaining issue lies in the intensity changes due to water absorption, scattering and the variations in altitude of the UHI.  For future work, it would be interesting to tackle the challenge of correcting the intensities of the UHI images to create constant illumination like reconstructions which could prove to be even more useful for scientific analysis.

\section{Conclusion}

The use of Underwater Hyperspectral Imaging (UHI) system has attracted a lot of attention from the scientific community because of the analysis perspective it offers.  Yet, as many UHI cameras work as push-broom scanners, most previous works have used it to create photo-mosaics based on a flat scene assumption and through the use of basic dead-reckoning navigation data, resulting in poorly accurate hyperspectral reconstructions.  In this paper we have proposed a new method for creating hyperspectral 3D reconstructions of underwater environments by cleverly fusing images acquired by an RGB camera and navigation data to the UHI images.  Creating such hyperspectral 3D reconstructions allows us to overcome the flat scene assumption which leads to geometrically distorted reconstructions.  We see this work as a step forward in the production of more reliable underwater hyperspectral material for scientists which will hopefully help in better understanding underwater environments and marine ecosystems.

\section{Acknowledgment}

This study is part of the joint Marha “Marine Habitats” (LIFE16 IPE FR001\_MARHA) and “EUMarineRobots” projects funded by the European Union.  The authors are thankful to the Ecotone company for lending the Underwater Hyperspectral Imaging camera used to acquire the data presented in this paper.

{\small
\bibliographystyle{ieee_fullname}
\bibliography{biblio}
}

\end{document}